# RAMCT: Novel Region-adaptive Multi-channel Tracker with Iterative Tikhonov Regularization for Thermal Infrared Tracking

Shang Zhang[1,2,3*], Yuke Hou[1,2,3], Guoqiang Gong[1,2,3], Ruoyan Xiong[1,2,3], and Yue Zhang[1,2,3]

[1] College of Computer and Information Technology, China Three Gorges University, Hubei, Yichang, 443002, China
[2] Hubei Province Engineering Technology Research Center for Construction Quality Testing Equipment, China Three Gorges University, Yichang 443002, China
[3] Hubei Key Laboratory of Intelligent Vision Based Monitoring for Hydroelectric Engineering, China Three Gorges University, Yichang 443002, China
zhangshang@ctgu.edu.cn

**Abstract.** Correlation filter (CF)-based trackers have gained significant attention for their computational efficiency in thermal infrared (TIR) target tracking. However, existing methods struggle with challenges such as low-resolution imagery, occlusion, background clutter, and target deformation, which severely impact tracking performance. To overcome these limitations, we propose RAMCT, a region-adaptive sparse correlation filter tracker that integrates multi-channel feature optimization with an adaptive regularization strategy. Firstly, we refine the CF learning process by introducing a spatially adaptive binary mask, which enforces sparsity in the target region while dynamically suppressing background interference. Secondly, we introduce generalized singular value decomposition (GSVD) and propose a novel GSVD-based region-adaptive iterative Tikhonov regularization method. This enables flexible and robust optimization across multiple feature channels, improving resilience to occlusion and background variations. Thirdly, we propose an online optimization strategy with dynamic discrepancy-based parameter adjustment. This mechanism facilitates real time adaptation to target and background variations, thereby improving tracking accuracy and robustness. Extensive experiments on LSOTB-TIR, PTB-TIR, VOT-TIR2015, and VOT-TIR2017 benchmarks demonstrate that RAMCT outperforms other state-of-the-art trackers in terms of accuracy and robustness.

## 1 Introduction

Thermal Infrared (TIR) target tracking has become a key research area in computer vision due to its extensive applications in low-visibility environments. TIR tracking utilizes advanced techniques from image processing, machine learning, and artificial intelligence, facilitating its adoption across diverse domains such as intelligent video surveillance, advanced driver-assistance systems, and industrial automation [1]. Its ability to operate effectively in poor lighting conditions and obscured environments has further increased its importance in these applications. Despite significant advancements, TIR tracking presents distinct challenges that are less common in visible-light tracking. In contrast to visible-light tracking, where large volumes of labeled data are readily available for pre-training robust models, TIR tracking encounters several inherent limitations. These include the relatively low feature richness of TIR imagery and considerable variations in target appearance, making it inherently more difficult to develop accurate and reliable tracking models. Moreover, TIR tracking is further hindered by challenges such as thermal crossover, intensity variation, background clutter, occlusions, and scale-variation [2, 3]. To address these issues, researchers have developed various methods and evaluated them using specialized TIR datasets, including LSOTB-TIR [4], PTB-TIR [5], VOT-TIR2015 [6], and VOT-TIR2017 [7]. However, despite these advancements, the lack of detailed and consistent feature representations in TIR imagery remains a major obstacle to develop robust TIR trackers.

---

[*]Corresponding author.

Correlation Filter (CF)-based methods have become a dominant approach in target tracking by utilizing the Fast Fourier Transform (FFT) to speed up convolution operations in the frequency domain. This computational efficiency makes CF-based trackers well-suited for real time applications [8]. However, despite their effectiveness, traditional CF models primarily depend on simple linear features and fixed regularization techniques, which reduce their ability to adapt to significant target deformations and complex background variations. To address these limitations, researchers have improved the discriminative capability of CF by integrating deep features and multi-channel feature fusion. For example, Li et al. [9] combined Histogram of Oriented Gradients (HOG) and color names (CN) features, effectively merging HOG gradients and CN colors to enhance tracking performance. Similarly, Gao et al. [10] utilized a pre-trained appearance network and a flow network to extract both appearance and motion features of thermal infrared targets, then integrated these with a structured support vector machine (S-SVM) to improve tracking accuracy.

While these methods enhance tracking performance, they often demand substantially greater computational resources, which restricts their feasibility for real time applications. Furthermore, CF-based methods face significant challenges in scenarios involving occlusion, background clutter, and target appearance variations. This is partly due to the lack of adaptive mechanisms for dynamically updating filter parameters in response to such changes. Additionally, conventional fixed-form regularization is often inadequate for effectively balancing sparsity and robustness. Both adaptive updating and flexible regularization are essential for maintaining stable tracking under partial occlusions and sudden appearance variations.

In practical TIR tracking scenarios, such as a pedestrian moving behind a heat source or entering a shadowed area, the target's thermal signature often fluctuates significantly. Under these conditions, static filter configurations tend to drift or fail, highlighting the importance of online optimization, which enables real time parameter updates in response to abrupt scene changes. For example, Wang et al. [11] proposed a variational online learning framework that incorporates confidence-aware updates and Kullback–Leibler divergence to improve both robustness and accuracy. Similarly, adaptive regularization plays a crucial role in applying context-sensitive constraints that adjust with the input, improving resilience to noise, occlusion, and thermal crossover. The effectiveness has been demonstrated by Dai et al. [12], who proposed Adaptive Spatially-Regularized Correlation Filters capable of dynamically adjusting spatial penalties based on the tracking context. Therefore, enhancing the adaptability of CF-based trackers through online optimization and adaptive regularization, while preserving computational efficiency, remains a key research challenge.

In this paper, we propose RAMCT, a sparse correlation filter tracker with online optimization, integrating multi-channel feature optimization with a GSVD-based region-adaptive iterative Tikhonov regularization. Extensive evaluations on LSOTB-TIR, VOT-TIR, and PTB-TIR benchmarks demonstrate that our RAMCT outperforms other state-of-the-art trackers in terms of accuracy and robustness.

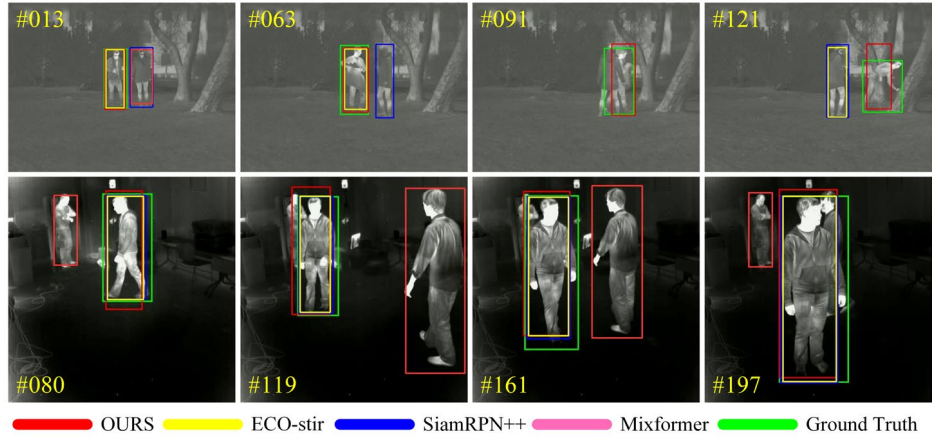

**Fig. 1.** Comparison between our proposed RAMCT and three state-of-the-art trackers.

Additionally, Fig. 1 illustrates that the tracking results of RAMCT closely match the Ground Truth labels of the target. The main contributions are summarized as follows:
- We propose a CF tracker that leverages Tikhonov regularization to improve noise resilience and ensure consistency across multi-channel features.
- We develop a spatially adaptive binary mask that enforces sparsity in the target region, effectively suppressing background interference while adapting to target deformation and occlusion.
- We propose a GSVD-based region-adaptive iterative Tikhonov regularization method to enable adaptive regularization across channels, enhancing robustness to handle occlusion and dynamic background changes.
- We introduce an online optimization strategy that dynamically adjusts regularization parameters in response to variations in target and background features, improving real time adaptability and computational efficiency.

## 2  Related Work

### 2.1  Tracking with Correlation Filter

Correlation filter (CF)-based tracking methods have attracted considerable attention due to their efficiency and high accuracy in visual tracking tasks. The foundational MOSSE tracker, introduced by Bolme et al. [13], marked the beginning of CF-based tracking and has since inspired a series of improvements. Henriques et al. [14] refined this method by integrating Histogram of Oriented Gradients (HOG) features, leading to the development of the Kernelized Correlation Filter (KCF) tracker. To better handle scale variations, Danelljan et al. [15] proposed a CF framework that employs a scale pyramid. Additionally, block-based strategies were introduced to improve robustness in the presence of occlusion. To handle long-term tracking challenges, Ma et al. [16] developed a framework capable of re-detection and re-tracking after tracking failures. This method separates the tracking task into translation and scale estimation while leveraging an online classifier for global re-detection. Furthermore, to minimize boundary effects during CF training, Danelljan et al. [17] introduced Spatially Regularized Correlation Filter (SRDCF), which penalizes correlation coefficients at the image edges. While these advances have substantially improved the accuracy of CF-based tracking, they have also increased computational demands, leading to slower tracking speeds.

With the rapid advancement of deep learning in computer vision, researchers have increasingly incorporated deep convolutional neural networks (CNNs) into the CF tracking framework to boost performance. For instance, Ma et al. [18] proposed a coarse-to-fine tracking model that integrates multi-layer convolutional features from CNNs into the CF structure. This approach outperforms traditional methods that rely on features from a single convolutional or fully connected layer. Similarly, Qi et al. [19] introduced an ensemble-based tracking framework that utilizes multi-layer convolutional features. Their method applies KCF to each convolutional layer and employs Hedge, an adaptive ensemble model, to fuse the outputs of multiple trackers. Danelljan et al. [20] enhanced SRDCF by incorporating deep convolutional features, leading to improved tracking performance. They further advanced this method by integrating multi-resolution convolutional features with a continuous interpolation model, which contributed to the development of more trackers such as C-COT [21] and ECO [22].

Building on the success of CF in TIR tracking, we propose a region-adaptive sparse correlation filter tracker, RAMCT, that combines multi-channel feature optimization with a GSVD-based region-adaptive iterative Tikhonov regularization. The primary innovation of this work lies in the integration of multi-channel features into the CF framework, and utilizing GSVD to enable adaptive regularization across feature channels. This method effectively overcomes the limitations of conventional CF methods in handling target deformation and partial occlusion under complex environmental conditions. In addition, the proposed online optimization strategy adaptively adjusts regularization parameters and efficiently solves subproblems, reducing computational cost while enhancing real time tracking performance.

### 2.2   Tikhonov Regularization

Tikhonov regularization, a widely used technique for solving ill-posed problems, has become a fundamental method in image processing tasks such as denoising, deblurring, and image reconstruction. It stabilizes optimization solutions by adding a regularization term, usually the squared Euclidean norm of the solution. This adjustment improves the conditioning of the problem [23].

In recent years, researchers have further refined and adapted Tikhonov regularization to handle various image processing challenges. Liu et al. [24] proposed a robust Tikhonov-based method for image denoising, which incorporates adaptive weighting to account for spatial variations in noise across image regions. Wang et al. [25] introduced a multi-scale strategy, showing that Tikhonov regularization can effectively manage the trade-off between preserving high-frequency details and reducing noise by assigning scale-specific regularization parameters. To improve the adaptability of Tikhonov regularization, researchers have developed adaptive techniques. Gazzola et al. [26] introduced an adaptive Tikhonov regularization method for dynamic image restoration, modifying regularization parameters in response to local image characteristics. Furthermore, recent studies have incorporated deep learning priors into Tikhonov regularization. Reichel [27] combined a neural network prior with Tikhonov regularization to enhance image reconstruction in data-scarce scenarios. Their method utilizes the advanced feature extraction capabilities of CNNs to refine the regularization process.

Building on these advancements, this paper presents an enhanced Tikhonov regularization technique that integrates iterative optimization with adaptive parameter adjustment. Specifically, we propose a GSVD-based region-adaptive iterative Tikhonov regularization method that imposes adaptive constraints across multiple feature channels. This approach enhances robustness to target deformation, occlusion, and background clutter, promoting stronger structural consistency during tracking. Furthermore, by incorporating an online optimization strategy guided by dynamic discrepancy principles, the regularization parameters are adjusted in real time, allowing the model to swiftly adapt to sudden changes in target appearance and environmental conditions. Collectively, these improvements lead to increased tracking accuracy, improved resilience in complex scenarios, and greater computational efficiency.

## 3   Methodology

### 3.1   Overview of RAMCT

In TIR tracking tasks, single-channel features often lack the robustness required to cope with complex environments. To handle this situation, we adopt multi-channel feature representations (*e.g.* HOG and CNN) to enhance target discriminability. These multi-channel features improve the capability of CF to effectively distinguish the target from the background [28]. However, the inclusion of multiple feature channels also increases the complexity of the optimization process and may introduce instability in the learned filters. To reinforce the stability and adaptability of the tracker, we further integrate Tikhonov regularization into the multi-channel CF framework. This regularization imposes smoothness constraints and mitigates overfitting by penalizing abrupt changes in the filter response, thereby improving resilience to noise, target deformation, and background clutter. As a result, the optimization objective is reformulated with adaptive regularization terms:

$$\arg\min_{x}\left\{\sum_{i=1}^{n}\alpha_i(\sum_{k=1}^{K_i}\theta_{ik}\parallel x-b_k\parallel^2)+\lambda\sum_{j=1}^{m}\beta_j(\sum_{l=1}^{L_j}\phi_{jl}\parallel L_{jl}(x-x^{(0)})\parallel^2)\\+\gamma(\sum_{r=1}^{R}\psi_r\parallel x-x^{(0)}\parallel^2)\right\} \quad (1)$$

In this formulation, $x$ denotes the correlation filter to be optimized, and $b_k$ represents the observed feature corresponding to the $k$-th component of the multi-channel representation. The weights $\alpha_i$ and $\theta_{ik}$ reflect the importance of the $i$-th channel and its components. $L_{jl}$ is a spatial regularization operator that promotes localized consistency, while $x^{(0)}$ is the initial filter estimate. The coefficients $\phi_{jl}$ and $\psi_r$ control the strength of spatial and temporal constraints, while $\lambda$ and $\gamma$ determine their overall contribution to the objective. This objective integrates three key components. The first term aligns the learned filter with multi-channel target features to enhance discriminability. The second term imposes spatial regularization, promoting smooth adaptation to deformations and background variation. The third term enforces temporal consistency by penalizing deviations from the previous filter estimate. Together, these terms improve robustness and adaptability in challenging TIR tracking scenarios.

To improve the clarity of the proposed RAMCT method, we illustrate its overall architecture and implementation process in Fig 2.

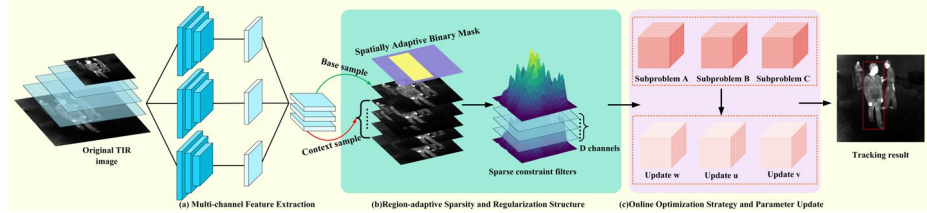

**Fig. 2.** The architecture and implementation process of RAMCT.

### 3.2   GSVD-based region-adaptive iterative Tikhonov regularization

Traditional CF often struggle with challenges such as target deformation and partial occlusion. To overcome these limitations, we introduce the RASCF tracker, which enhances conventional filtering techniques by integrating region-adaptive sparsity and context-aware regularization. The core innovation of RASCF lies in a spatially adaptive binary mask that concentrates response of the filter on the target region while effectively suppressing irrelevant background features. Unlike traditional CF that apply a uniform filtering response across the entire search area, our RASCF dynamically modifies the mask based on the position, thereby improving robustness against background clutter and occlusions. In addition, the model incorporates context-aware regularization, which refines the filter coefficients according to the surrounding context of the target. This allows the filter to better adapt to changes in scale, rotation, and partial occlusion. By leveraging this context-driven adaptability, our RASCF achieves enhanced resilience and accuracy, making it effective in real-world tracking scenarios.

To further improve model performance, we incorporate the proposed GSVD into the optimization framework. Tikhonov regularization enhances stability by reducing the risk of overfitting, while GSVD boosts computational efficiency by decomposing filter coefficients into distinct subspaces [29, 30]. The iterative nature of Tikhonov regularization allows for the progressive refinement of the filter coefficients, resulting in greater tracking accuracy and robustness. By combining region-adaptive sparsity, context-aware regularization, and advanced optimization techniques, our RASCF demonstrates reliable tracking performance under challenging conditions, such as occlusion, deformation, and background clutter.

The optimization problem for the proposed RASCF is formulated as follows:

$$\arg\min_{h} \left\{ \sum_{i=1}^{p} \sigma_i (\sum_{k=1}^{K_i} \rho_{ik} \parallel \Sigma_{ik} h - r_k^{(k)} \parallel^2) + \sum_{j=1}^{q} \mu_j^{(k)} (\sum_{l=1}^{L_j} \tau_{jl} \parallel \Lambda_{jl} h \parallel^2) + \theta (\sum_{r=1}^{R} \eta_r \parallel h - h^{(0)} \parallel^2) \right\} \quad (2)$$

where $h$ denotes the filter being optimized in the transformed domain, such as the frequency space or GSVD subspace. The matrices $\Sigma_{ik}$ and $\Lambda_{jl}$ are diagonal components resulting from GSVD decomposition applied to the feature and regularization terms, respectively. The symbol $r_k^{(k)}$ refers to the ideal response map corresponding to the $k$-th feature component. The coefficient $\sigma_i$ controls the contribution of each feature channel in the fidelity term, while $\mu_j^{(k)}$ adjusts the strength of structure-aware regularization.

### 3.3  Dynamic Online Strategy for Efficient Filter Optimization

To minimize computational costs and improve adaptability, we propose an online optimization strategy that employs a dynamic discrepancy principle to automatically modify regularization parameters based on variations in target and background features. To further enhance the efficiency of the iterative Tikhonov regularization process, we incorporate refinements from Subproblems A, B, and C, integrating them with GSVD and zero-finding to help both high computational efficiency and precise estimation.

**Subproblem A.** In each iteration, we start by fixing $u$ and $v$, then introduce a relaxation formulation to update $w$. The objective function is reformulated as follows:

$$\arg\min_{w} \left\{ \frac{1}{2} \sum_{k=1}^{K} \alpha_k (\sum_{s=1}^{S_k} \omega_{ks} \parallel A_{ks} w - b_{ks} \parallel^2) + \frac{\lambda}{2} \sum_{j=1}^{m} \beta_j (\sum_{l=1}^{L_j} \phi_{jl} \parallel L_{jl} (w - w^{(0)}) \parallel^2) + \frac{\mu}{2} \parallel w - u \parallel^2 + v \parallel w \parallel^2 \right\} \quad (3)$$

where $A_{ks}$ denotes the feature transformation matrix for the $s$-th sample of the $k$-th component, and $b_{ks}$ is the corresponding target response. The scalar $\omega_{ks}$ modulates the relative importance of each sample, and $S_k$ is the number of samples for component $k$. This formulation accounts for both multi-sample matching accuracy and prior-based regularization. The parameter $\mu$ is a relaxation penalty that controls the similarity between $w$ and the auxiliary variable $u$, while $\nu$ imposes an $l_2$ regularization constraint on $w$ to prevent overfitting. By applying Tikhonov regularization and GSVD decomposition, we reformulate the problem as a generalized singular value decomposition problem:

$$w^{(k+1)} = \frac{\sum_{k=1}^{K} \sum_{s=1}^{S_k} \alpha_k \omega_{ks} \Sigma_{ks}^\top b_{ks} + \mu \sum_{j=1}^{q} \sum_{l=1}^{L_j} \phi_{jl} \Lambda_{jl}^\top \Lambda_{jl} u}{\sum_{i=1}^{p} \sum_{k=1}^{K_i} \rho_{ik} \Sigma_{ik}^\top \Sigma_{ik} + \sum_{j=1}^{q} \sum_{l=1}^{L_j} \tau_{jl} \mu_j \Lambda_{jl}^\top \Lambda_{jl} + \nu I} \quad (4)$$

where, $I$ is the identity matrix. The numerator reflects data fidelity and prior information, while the denominator combines data and regularization constraints.

**Subproblem B.** Solving for $u$ given $w$ and $v$. With $w$ and $v$ held fixed, the optimization problem for $u$ is formulated as follows:

$$\arg\min_{u} \left\{ \frac{\gamma}{2} \sum_{i=1}^{p} \left( \sum_{s=1}^{S_i} \chi_{is} \| u - w_{is} \|^2 \right) + \frac{\beta}{2} \sum_{j=1}^{q} \left( \sum_{t=1}^{T_j} \kappa_{jt} \| u - v_{jt} \|^2 \right) + \xi \left( \sum_{r=1}^{R} \psi_r \| u - u^{(0)} \|^2 \right) \right\} \quad (5)$$

where the scalars $\chi_{is}$ and $\kappa_{jt}$ represent confidence weights assigned to the current estimates $w_{is}$ and $v_{jt}$, respectively, while $\xi$ and $\psi_r$ regulate proximity to the prior solution $u^{(0)}$.

Using a zero-finder algorithm and introducing $\xi$ to control the proximity to the reference solution $u^{(0)}$, we derive the closed-form solution:

$$u^{(k+1)} = \frac{\gamma \sum_{i=1}^{p} \sum_{s=1}^{S_i} \chi_{is} w_{is} + \beta \sum_{j=1}^{q} \sum_{t=1}^{T_j} \kappa_{jt} v_{jt} + \xi \sum_{r=1}^{R} \psi_r u^{(0)}}{\gamma \sum_{i=1}^{p} S_i + \beta \sum_{j=1}^{q} T_j + \xi R} \quad (6)$$

This update promotes smooth evolution of $u$ by integrating current estimates with historical prior, enhancing robustness to abrupt changes.

**Subproblem C.** Updating $v$ using the Lagrange multiplier adjustment. To accommodate a dynamic background, the Lagrange multiplier update for $v$ follows a discrepancy-based adjustment criterion, expressed as:

$$v^{(k+1)} = v^{(k)} + \rho \left( \sum_{s=1}^{S} \tau_s (w^{(k+1)} - u^{(k+1)}) \right) + \eta \left( \sum_{r=1}^{R} \psi_r (v^{(0)} - v) \right) \quad (7)$$

where $\rho$ penalizes inconsistency between $w$ and $u$, while $\eta$ balances responsiveness and regularity by limiting deviation from the prior $v^{(0)}$. To adaptively regulate the penalty strength, the penalty factor $\rho$ is updated using the following rule:

$$\rho^{(k+1)} = \min(\rho_{max}, \rho^{(k)} + \delta \sum_{s=1}^{S} \tau_s \parallel w^{(k+1)} - u^{(k+1)} \parallel) \tag{8}$$

This adjustment improves the adaptability of model to dynamic background conditions and enhances its robustness to noise. To systematically implement the proposed optimization framework, we design a structured iterative algorithm. As shown in Algorithm 1, it details the step-by-step procedure for solving Subproblems A, B, and C, along with the penalty term updates and convergence checks. In each iteration, the variables $w$, $u$, and $v$, are progressively refined, while the regularization parameters are dynamically updated to ensure both stability and accuracy in varying conditions.

**Algorithm 1:** Online Optimization with Dynamic Discrepancy Adjustment

**Input:**
    Data matrix $A \in \mathbb{R}^{n \times m}$
    Initial guess, $u^{(0)}, v^{(0)}$
    Regularization parameters $\alpha_k, \beta_j, \lambda, \mu, \nu, \xi, \rho_{max}$
    Subproblem-specific parameters

01: Initialize $k \leftarrow 0$
02: repeat:
03:    Update $w$ by solving Subproblem A (Eq. 4)
04:    Update $u$ by solving Subproblem B given $w$ and $v$ (Eq. 6)
05:    Update $v$ via Lagrange multiplier adjustment (Eq. 7)
06:    Update penalty factor $\rho$ (Eq. 8)
07:    Check convergence (*e.g.*, relative error or max iterations)
08:    $k \leftarrow k + 1$
09: until convergence
10: return $w, u, v$

**Output:**
    Optimized parameters $w, u, v$

### 3.4 Auxiliary Variable Optimization with Iterative Refinement

To further minimize the high computational cost associated with pseudo-inverse operations, we reformulate the objective function by introducing auxiliary variables:

$$\arg\min_{w,u,v} \left\{ \sum_{i=1}^{n} \alpha_i (\sum_{k=1}^{K_l} \theta_{ik} \parallel A_{ik}w - b \parallel^2) + \parallel w - u \parallel^2 + \parallel u - v \parallel^2 \right. \\ \left. + \sum_{j=1}^{m} \beta_j (\sum_{l=1}^{L_j} \phi_{jl} \parallel L_j(w - w^{(0)}) \parallel^2) + \xi(\sum_{r=1}^{R} \psi_r \parallel w - w^{(0)} \parallel^2) \right\} \tag{9}$$

where denotes $A_{ik}$ the transformation matrix applied to the $k$-th feature in the $i$-th channel, mapping it into the response domain; $b$ is the desired target response.

This reformulation improves computational efficiency by decomposing the original problem into smaller, tractable subproblems that can be solved iteratively. The use of auxiliary variables simplifies the optimization, minimizes the cost of expensive pseudo-inverse calculations, and reduces the overall computational burden. To update $w$, $u$, and $v$ accordingly, we adopt an alternating minimization scheme as follows:

**Update $w$.** Holding $u$ and $v$ fixed, the optimization objective for $w$ is given by:

$$\arg\min_{w} \left\{ \frac{1}{2} \sum_{i=1}^{p} \sum_{k=1}^{K_i} \rho_{ik} \| A_{ik}w - b \|^2 + \frac{1}{2} \| w - u \|^2 + \frac{1}{2} \sum_{j=1}^{q} \sum_{l=1}^{L_j} \tau_{jl} \mu_j \| L_j(w - w^{(0)}) \|^2 \right\} \quad (10)$$

Taking the partial derivative with respect to $w$ and setting it to zero yields:

$$\left( \sum_{i=1}^{p} \sum_{k=1}^{K_i} \rho_{ik} A_{ik}^\top A_{ik} + \sum_{j=1}^{q} \sum_{l=1}^{L_j} \tau_{jl} \mu_j L_{jl}^\top L_{jl} + I \right) w = \sum_{i=1}^{p} \sum_{k=1}^{K_i} \rho_{ik} A_{ik}^\top b + u \quad (11)$$

Let:

$$P = \sum_{i=1}^{p} \sum_{k=1}^{K_i} \rho_{ik} A_{ik}^\top A_{ik} + \sum_{j=1}^{q} \sum_{l=1}^{L_j} \tau_{jl} \mu_j L_{jl}^\top L_{jl} \quad (12)$$

Thus, the equation simplifies to:

$$(P + I)w = \sum_{i=1}^{p} \sum_{k=1}^{K_i} \rho_{ik} A_{ik}^\top b + u \quad (13)$$

Solving for $w$, we derive:

$$w = (P + I)^{-1} \left( \sum_{i=1}^{p} \sum_{k=1}^{K_i} \rho_{ik} A_{ik}^\top b + u \right) \quad (14)$$

**Update $u$.** The variable $u$ is updated by minimizing its deviation from the current filter estimate $w$, the dual variable $v$, and the previous filter $u^{(0)}$:

$$\arg\min_{u} \left\{ \frac{\gamma}{2} \sum_{i=1}^{p} \sum_{s=1}^{S_i} \| u - w_{is} \|^2 + \frac{\beta}{2} \sum_{j=1}^{q} \sum_{t=1}^{T_j} \kappa_{jt} \| u - v_{jt} \|^2 + \frac{\xi}{2} \| u - u^{(0)} \|^2 \right\} \quad (15)$$

Taking the partial derivative with respect to $u$ and setting it to zero yields:

$$(\gamma \sum_{i=1}^{p} S_i + \beta \sum_{j=1}^{q} T_j + \xi)u = \gamma \sum_{i=1}^{p} \sum_{s=1}^{S_i} \chi_{is} w_{is} + \beta \sum_{j=1}^{q} \sum_{t=1}^{T_j} \kappa_{jt} v_{jt} + \xi u^{(0)} \quad (16)$$

Simplifying, we obtain:

$$u = \frac{\gamma \sum_{i=1}^{p} \sum_{s=1}^{S_i} \chi_{is} w_{is} + \beta \sum_{j=1}^{q} \sum_{t=1}^{T_j} \kappa_{jt} v_{jt} + \xi u^{(0)}}{\gamma \sum_{i=1}^{p} S_i + \beta \sum_{j=1}^{q} T_j + \xi} \quad (17)$$

**Algorithm 2:** Auxiliary Variable Optimization with GSVD and Iterative Refinement

**Input:**
  Feature matrix $A_{ik} \in \mathbb{R}^{n \times m}$
  Initial variables $w^{(0)}, u^{(0)}, v^{(0)}$
  Parameters: $\alpha_i, \beta_j, \gamma, \xi, \zeta, \lambda, \rho$
  Tuning parameters: $\{\theta_{ik}, \phi_{jl}, \psi_r\}$

01: Initialize $k \leftarrow 1$
02: **while** convergence condition (Eq. 9) not satisfied **do**
03:     Update $w^{(k+1)}$ using Eq.(12) and Eq.(14) with $u^{(k)}, v^{(k)}$
04:     Update $u^{(k+1)}$ using Eq.(17) with $w^{(k+1)}, v^{(k)}$
05:     Update $v^{(k+1)}$ using Eq.(20) with $w^{(k+1)}, u^{(k+1)}$
06:     (Optional) Refine $v^{(k+1)}$ using Eq. (21)
07:     **if** convergence criterion is met **then**
08:         break
09:     end if
10:     $k \leftarrow k + 1$
11: end while
12: Return $w^{(k)}, u^{(k)}, v^{(k)}$

**Output:**
  Optimized variables: $w, u, v$

**Update $v$.** The dual variable $v$ is then updated to balance proximity to $u$ and its previous value $v^{(0)}$, with $\zeta$ controlling the strength of temporal regularization:

$$\arg\min_{v} \left\{ \frac{1}{2} \| v - u \|^2 + \frac{\zeta}{2} \| v - v^{(0)} \|^2 \right\} \quad (18)$$

Taking the partial derivative with respect to $v$ and setting it to zero, we obtain:

$$(I + \zeta I)v = u + \zeta v^{(0)} \quad (19)$$

Solving for $v$, we derive:

$$v = (I + \zeta I)^{-1}(u + \zeta v^{(0)}) \quad (20)$$

Applying the matrix inversion lemma, if $I + \zeta I$ is diagonal, the inversion process:

$$v^{(k+1)} = v^{(k)} + \rho(\sum_{s=1}^{S} \tau_s(w^{(k+1)} - u^{(k+1)})) + \eta(\sum_{r=1}^{R} \psi_r(v^{(0)} - v^{(k)})) \qquad (21)$$

This iterative refinement process improves convergence speed while maintaining numerical stability. By leveraging the GSVD technique, each update step incorporates structured regularization efficiently, and enhances the robustness of auxiliary variable estimation. The steps of this optimization method are shown in Algorithm 2.

Fig.3 illustrates the overall iterative structure of the RAMCT method, combining alternating subproblem updates with auxiliary refinement and adaptive regularization.

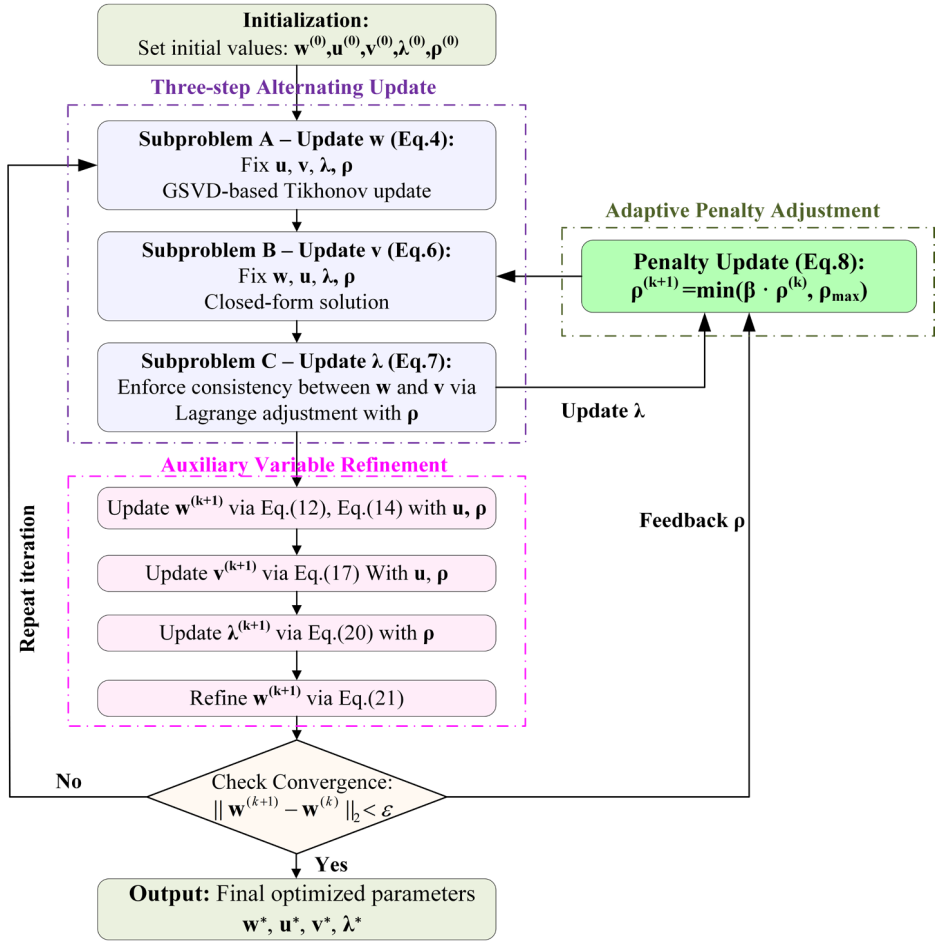

**Fig. 3.** Schematic diagram of the iterative update process in the proposed RAMCT method.

RAMCT: Sparse Correlation Filter Tracker for Thermal Infrared Target Tracking    13## 3.5 Convergence and Stability Analysis

To assess the effectiveness of the proposed optimization algorithm, we define error criteria and convergence metrics to quantify both convergence speed and stability.

The loss function at the $k$-th iteration is formulated as:

$$Loss^{(k)} = \sum_{i=1}^{p} \sum_{k=1}^{K_i} \theta_{ik} \| A_{ik} w^{(k)} - b \|^2 + \sum_{j=1}^{m} \beta_j \sum_{l=1}^{L_j} \varphi_{jl} \| L_j(w^{(k)} - w^{(0)}) \|^2 \\ + \| w^{(k)} - u^{(k)} \|^2 + \| u^{(k)} - v^{(k)} \|^2 + \zeta \sum_{r=1}^{R} \psi_r \| v^{(k)} - v^{(0)} \|^2 \quad (22)$$

This loss function captures overall model performance by aggregating the reconstruction error, regularization penalties, and variable differences across iterations.

The stability of the variable $v$ is measured using the Lagrange multiplier update:

$$\Delta v^{(k+1)} = v^{(k+1)} - v^{(k)} + \eta (\sum_{r=1}^{R} \psi_r (v^{(0)} - v)) \quad (23)$$

This metric quantifies how $v$ evolves in relation to its initial reference $v^{(0)}$, helping consistent updates throughout the iterations. To optimize regularization parameters and monitor convergence, we use the Mean Squared Error (MSE):

$$MSE = \frac{1}{N} \sum_{i=1}^{N} \gamma_i \| x^{(k)} - x^{true} \|^2 \quad (24)$$

where $x^{(k)}$ denotes the solution estimate at iteration $k$, while $x^{true}$ represents the ground truth. MSE serves as an objective indicator of tracking accuracy, guiding parameter tuning during optimization.

# 4    Experiment

In this section, we conduct extensive experiments to assess the effectiveness and robustness of the proposed RAMCT. The evaluation is performed on the LSOTB-TIR, PTB-TIR, VOT-TIR2015, and VOT-TIR2017 benchmarks.

## 4.1    Implementation Details

**Datasets and Evaluation Criteria.** LSOTB-TIR is the largest and most comprehensive benchmark for thermal infrared (TIR) tracking. It comprises 1,400 TIR video sequences, spanning 600K frames and covering 47 distinct object categories. The dataset features 4 scenario attributes and 12 challenge attributes, offering a robust framework for evaluating TIR tracking algorithms. PTB-TIR is a specialized benchmark designed for pedestrian tracking in thermal infrared environments. It consists of 60 manually annotated pedestrian sequences with a total of 30K frames. The dataset presents 9 distinct tracking challenges, including scale variation, background clutter, and fast motion, providing a comprehensive benchmark for assessing the robustness of TIR pedestrian trackers.VOT-TIR2015 and VOT-TIR2017, released by the Visual Object Tracking (VOT) committee, are widely recognized authoritative benchmarks in the TIR tracking domain. They encompass a range of challenging scenarios, serving as a rigorous standard for evaluating tracker performance in terms of accuracy, robustness, and adaptability across diverse conditions [6], [7]. Following [4] and [5], we evaluate the tracker on the LSOTB-TIR and PTB-TIR datasets using the One Pass Evaluation (OPE) protocol. The performance of RAMCT is quantitatively analyzed from multiple perspectives. For the PTB-TIR dataset, we use Success (S) and Precision (P) rates to assess tracking performance. However, for the LSOTB-TIR dataset, we additionally introduce Normalized Precision (NP) as an evaluation metric.

**Experimental Platform and Tracker Parameter Setting.** The proposed RAMCT is implemented in Python 3.7.1 on an Ubuntu 18.04 computer equipped with an Intel i7-12700 CPU @ 2.10 GHz, 32GB of RAM, NVIDIA RTX 3070 GPU 8GB. We employ CUDA 12.4 and CUDNN 8.9.6 to accelerate the filter training process. Table 1 summarizes the key parameters used in RAMCT, including their descriptions and values.

**Table 1.** Tracker parameter setting.

| Parameter | Description | Value |
|---|---|---|
| $\lambda$ | Regularization weight for structural prior | 0.6 |
| $\mu$ | Relaxation penalty between $w$ and $u$ | 4.0 |
| $\rho$ | Discrepancy penalty factor (initial) | 0.1, max=1.0 |
| $\eta$ | Penalty on deviation from initial $v^{(0)}$ | 0.2 |
| $\nu$ | $l_2$ regularization coefficient | 0.001 |

### 4.2    Ablation Experiment

In this section, we conduct internal comparison experiments on the PTB-TIR and LSOTB-TIR benchmarks to evaluate the contribution of each component within the proposed RAMCT tracker. First, we use ATOM [31] as the baseline tracker and compare it with ATOM_GRAITR, which integrates the GSVD-based Region-Adaptive Iterative Tikhonov Regularization (GRAITR), to assess its impact. Second, we compare ATOM with ATOM_DOS, which incorporates the Dynamic Online Strategy (DOS), to evaluate its effectiveness. These comparisons help isolate and quantify the performance gains introduced by each individual module. The results of the ablation study are summarized in Table 2.

As shown in Table 2, when the dynamic online strategy is integrated independently into the ATOM framework, the PTB-TIR benchmark shows improvements of 12.9% in Precision and 5.1% in Success rate. Additionally, the LSOTB-TIR benchmark records gains of 14.1% in Precision, 9.7% in Normalized Precision, and 7.9% in Success rate. These results indicate that DOS effectively enhances multi-channel feature extraction and introduces regional sparsity into the correlation filter, resulting in more accurate response maps and improved localization performance, particularly under background clutter and occlusion. Similarly, incorporating the GSVD-based region-adaptive iterative Tikhonov regularization alone leads to even greater improvements. Specifically, Precision increases by 15.3% on PTB-TIR and 16.4% on LSOTB-TIR, while Success rate improves by 9.3% and 9.7%, respectively. In summary, the ablation study confirms that each component contributes meaningfully to the overall performance enhancement of the proposed RAMCT tracker.

Table 2. Ablation analysis on PTB-TIR and LSOTB-TIR benchmark.

| Model | Component | | PTB-TIR | | LSOTB-TIR | | | FPS |
|---|---|---|---|---|---|---|---|---|
| | DOS | GRAITR | Precision (P/%) | Success (S/%) | Precision (P/%) | Norm. Pre. (NP/%) | Success (S/%) | |
| ATOM (baseline) | | | 57.4 | 51.1 | 59.2 | 53.8 | 49.2 | **30.0** |
| ATOM_DOS | √ | | 70.3 | 56.2 | 73.3 | 63.5 | 57.1 | 26.7 |
| ATOM_GRAITR | | √ | 72.7 | 60.4 | 75.6 | 60.7 | 58.9 | 23.5 |
| RAMCT (ours) | √ | √ | **80.3** | **62.7** | **82.5** | **68.1** | **65.2** | 13.0 |

### 4.3   Performance Comparison with State-of-the-arts

**Results on Attributes in LSOTB-TIR and PTB-TIR.** To further evaluate the robustness of the proposed RAMCT tracker under diverse challenging conditions, we conducted comprehensive experiments comparing it against several trackers using 12 attributes defined in LSOTB-TIR and PTB-TIR. These attributes include deformation (DF), scale variation (SV), occlusion (OCC), fast motion (FM), background clutter (BC), motion blur (MB), intensity variation (IV), low resolution (LR), thermal crossover (TC), out-of-view (OV), aspect ratio change (ARC), and distractors (DI).

Table 3. Attributes analysis on LSOTB-TIR benchmark.

| Attributes Type | Attributes | Tracker (P/% ↑   NP/% ↑   S/% ↑) | | | | | |
|---|---|---|---|---|---|---|---|
| | | MDNet [32] | DFG [33] | TransT [34] | ECO-stir [35] | VITAL [36] | RAMCT(ours) |
| Challenge | DF | 75.9/63.7/49.1 | 69.9/61.1/58.4 | 66.4/56.3/47.9 | 79.3/65.8/52.7 | 77.5/64.9/50.8 | **86.6/73.2/64.3** |
| | OCC | 70.5/62.3/45.6 | 71.9/68.9/54.2 | 60.9/55.2/48.3 | 78.3/**75.4**/51.9 | 69.3/60.5/46.2 | **85.8**/68.7/**57.5** |
| | SV | 85.6/72.8/33.5 | **93.6**/65.9/52.9 | 70.1/66.0/51.8 | 83.5/74.9/65.7 | 84.9/**77.8/66.2** | 84.3/73.5/50.8 |
| | BC | 74.3/62.8/61.7 | 68.9/67.3/55.1 | **81.2/72.5/62.8** | 73.9/67.0/48.5 | 70.1/65.9/56.7 | 75.8/63.4/60.2 |
| | LR | 85.1/**79.2**/65.5 | 86.3/75.3/**69.7** | 78.2/68.4/49.0 | 83.3/64.8/44.9 | 81.7/66.9/47.5 | **90.5**/76.1/67.3 |
| | FM | 79.5/71.4/63.8 | 80.4/68.0/61.6 | 74.9/64.8/47.3 | 75.3/64.9/57.9 | 76.7/63.2/58.1 | **85.2/74.8/66.5** |
| | MB | 77.2/68.7/59.5 | 78.9/67.5/57.8 | 69.1/55.6/49.8 | 76.1/61.9/56.4 | 74.6/60.4/54.9 | **85.0/75.3/65.2** |
| | OV | 70.3/64.4/57.1 | 69.7/60.3/49.0 | 68.5/64.3/47.6 | 69.2/63.5/58.0 | **79.8**/63.7/56.4 | 72.6/**70.2/65.3** |
| | IV | 78.5/76.3/**67.9** | 90.7/63.4/56.5 | 67.3/60.7/52.9 | 89.0/**84.5**/61.8 | 80.4/74.6/59.2 | 85.2/72.9/62.0 |
| | TC | 67.6/59.7/41.2 | 69.3/64.2/42.9 | 58.4/53.9/**48.8** | 59.1/48.5/39.7 | 61.2/50.7/38.6 | **73.8/68.4**/40.5 |
| | ARC | 73.1/61.4/50.9 | 77.9/**71.8**/56.7 | 68.8/59.7/45.9 | **82.7**/57.9/51.8 | 71.5/58.4/53.1 | 74.5/65.7/**59.3** |
| | DI | 75.2/64.3/56.5 | 76.8/65.7/54.2 | 66.5/62.2/44.6 | 72.1/63.5/48.3 | 73.4/61.8/49.7 | **81.3/77.1/59.1** |
| | ALL | 76.1/67.3/54.4 | 77.9/66.6/55.8 | 69.2/61.6/49.7 | 76.8/66.1/53.1 | 75.1/64.1/53.1 | **81.7/71.6/59.8** |

As illustrated in Table 3, the proposed RAMCT achieves the highest overall scores on the LSOTB-TIR benchmark, with 81.7% in Precision, 71.6% in Normalized Precision, and 59.8% in Success rate. It surpasses other trackers across a wide range of attributes, ranking first in Precision for six attributes and in Success rate for five. Notably, RAMCT demonstrates strong performance in some challenging scenarios, including deformation, occlusion, and thermal crossover. Although ECO-stir achieves the best Precision in the ARC attribute, RAMCT still delivers better overall results, with respective gains of 5.1% in Precision, 4.8% in Normalized Precision, and 6.7% in Success rate.

Table 4. Attributes analysis on PTB-TIR benchmark.

| Attributes Type | Attributes | Tracker (P/% ↑ S/% ↑) | | | | | |
|---|---|---|---|---|---|---|---|
| | | ECO-stir | Mixformer | ATOM | VITAL | SiamRPN++ | RAMCT(ours) |
| Challenge | DF | 82.1/63.7 | 75.8/54.9 | 80.9/58.4 | 72.1/48.6 | 83.6/64.2 | **86.3/69.5** |
| | OCC | 81.7/54.1 | 73.5/58.3 | 80.1/64.5 | 77.2/66.9 | 84.3/49.5 | **87.0/70.8** |
| | SV | 77.1/45.9 | 69.2/57.6 | **83.1/72.5** | 62.0/50.3 | 72.4/56.9 | 73.0/60.8 |
| | BC | 81.0/59.2 | 79.5/56.2 | 73.0/55.1 | 62.4/45.5 | 71.1/53.0 | **86.6/65.9** |
| | LR | **88.6**/56.2 | 66.2/53.8 | 79.3/60.4 | 69.4/50.6 | 78.8/**68.3** | 77.1/61.3 |
| | FM | 80.5/60.9 | 87.0/**79.3** | 84.6/68.7 | 82.3/49.8 | 75.3/62.4 | **91.4**/61.9 |
| | MB | 81.7/61.8 | 78.0/**69.5** | 79.8/52.7 | 69.9/55.2 | 72.1/57.7 | **87.3**/60.3 |
| | OV | 73.7/58.1 | 77.2/55.8 | 66.6/63.3 | 78.3/45.9 | **87.2**/58.5 | 69.4/**73.1** |
| | IV | 73.5/45.3 | 82.2/59.9 | 71.4/68.0 | **88.3/75.0** | 79.6/56.1 | 81.9/67.3 |
| | TC | 82.5/68.9 | 74.3/60.7 | 80.1/55.4 | 83.2/69.3 | 76.4/61.8 | **90.5/78.0** |
| | ARC | 77.9/60.2 | **85.1/71.9** | 83.6/68.3 | 68.7/52.4 | 74.5/59.6 | 79.4/62.1 |
| | DI | 79.2/61.5 | 84.6/63.8 | 76.3/57.9 | 81.4/58.2 | 72.8/55.7 | **89.7/66.4** |
| | ALL | 80.0/58.0 | 77.7/61.8 | 78.2/62.1 | 74.6/55.6 | 77.3/58.6 | **83.3/66.5** |

Table 4 compares the performance of the proposed RAMCT tracker with other state-of-the-art trackers on the PTB-TIR benchmark. Our RAMCT demonstrates better tracking capabilities, achieving the highest overall Precision and Success rate at 83.3% and 66.5%, respectively. These results underscore its robustness across diverse and challenging tracking conditions. While MixFormer [37] records the highest Success rate in three attributes (ARC, MB, and FM), our RAMCT outperforms it overall, with improvements of 5.6% in Precision and 4.7% in Success rate. Notably, RAMCT shows particular strength in attributes such as DF, OCC, LR, TC, and DI. Additionally, although SiamRPN++ [38] performs competitively in OCC and OV scenarios, our RAMCT still delivers better results, with gains of 5.6% in Precision and 8.2% in Success rate.

**Results on VOT-TIR2015 and VOT-TIR2017.** Table 5 highlights the performance of the proposed RAMCT tracker, which surpasses previous state-of-the-art trackers on both VOT-TIR2015 and VOT-TIR2017 benchmarks. Our RAMCT achieves the highest accuracy scores of 0.86 on VOT-TIR2015 and 0.77 on VOT-TIR2017, outperforming SiamRPN++ and ATOM. It also records the best Expected Average Overlap (EAO) scores, with 0.334 and 0.336 on the respective benchmarks, indicating its strong overall tracking effectiveness. In addition, RAMCT achieves competitive robustness scores of 2.09 and 2.13, reflecting its resilience in challenging tracking scenarios. Compared to the CF-based tracker ECO-MM, RAMCT delivers notable improvements in accuracy, with gains of 0.14 on VOT-TIR2015 and 0.12 on VOT-TIR2017.

Table 5. Performance Comparison on VOT-TIR 2015 and 2017 benchmark.

| Method | Trackers | VOT-TIR 2015 | | | VOT-TIR 2017 | | | FPS |
|---|---|---|---|---|---|---|---|---|
| | | EAO↑ | Acc↑ | Rob↓ | EAO↑ | Acc↑ | Rob↓ | |
| Deep learning | CREST (2017) | 0.258 | 0.62 | 3.11 | 0.252 | 0.59 | 3.26 | 0.6 |
| | VITAL (2018) | 0.289 | 0.63 | 2.18 | 0.272 | 0.64 | 2.68 | 4.7 |
| | DeepSTRCF (2018) | 0.257 | 0.63 | 2.93 | 0.262 | 0.62 | 3.32 | 5.5 |
| | DiMP (2019) | 0.330 | 0.69 | 2.23 | 0.328 | 0.66 | 2.38 | 40.0 |
| | Ocean (2020) | 0.339 | 0.70 | 2.43 | 0.320 | 0.68 | 2.83 | 25.0 |
| | UDCT (2022) | **0.680** | 0.67 | 0.88 | 0.321 | 0.52 | 0.90 | - |
| Siamese Network | SiamFC (2016) | 0.219 | 0.60 | 4.10 | 0.188 | 0.50 | **0.59** | 66.9 |
| | DaSiamRPN (2018) | 0.311 | 0.67 | 2.33 | 0.258 | 0.62 | 2.90 | 110.0 |
| | SiamRPN (2018) | 0.267 | 0.63 | 2.53 | 0.242 | 0.60 | 3.19 | **160.0** |
| | SiamRPN++ (2019) | 0.313 | 0.74 | 2.25 | 0.296 | 0.69 | 2.63 | 46.0 |
| | HSSNet (2019) | 0.311 | 0.67 | 2.53 | 0.262 | 0.58 | 3.33 | 10.0 |
| | MLSSNet (2020) | 0.329 | 0.57 | 2.42 | 0.286 | 0.56 | 3.11 | 18.0 |
| Transformer | TransT (2021) | 0.287 | 0.77 | 2.75 | 0.290 | 0.71 | 0.69 | 50.0 |
| | CorrFormer (2023) | 0.269 | 0.71 | **0.56** | 0.262 | 0.66 | 1.23 | 37.0 |
| | DFG (2024) | 0.329 | 0.78 | 2.41 | 0.304 | 0.74 | 2.63 | - |
| Correlation filter | SRDCF (2015) | 0.225 | 0.62 | 3.06 | 0.197 | 0.59 | 3.84 | 12.3 |
| | HDT (2016) | 0.188 | 0.53 | 5.22 | 0.196 | 0.51 | 4.93 | 10.6 |
| | ECO-deep (2017) | 0.286 | 0.64 | 2.36 | 0.267 | 0.61 | 2.73 | 16.3 |
| | MCFTS (2017) | 0.218 | 0.59 | 4.12 | 0.193 | 0.55 | 4.72 | 4.7 |
| | MCCT (2018) | 0.250 | 0.67 | 3.34 | 0.270 | 0.53 | 1.76 | - |
| | ATOM (2019) | 0.331 | 0.65 | 2.24 | 0.290 | 0.61 | 2.43 | 30.0 |
| | ECO-MM (2022) | 0.303 | 0.72 | 2.44 | 0.291 | 0.65 | 2.31 | - |
| | ECO_LS (2023) | 0.319 | 0.64 | 0.82 | 0.302 | 0.55 | 0.93 | - |
| | ECOHG_LS (2023) | 0.270 | 0.60 | 0.92 | 0.251 | 0.49 | 1.26 | - |
| | RAMCT (ours) | 0.334 | **0.86** | 2.09 | **0.336** | **0.77** | 2.13 | 13.0 |

### 4.4 Visualized Comparison Results

We provide visual tracking results of the proposed RAMCT and several representative trackers (MCCT, AMFT, MixFormer) across six challenging sequences selected from the LSOTB-TIR and PTB-TIR benchmarks. As shown in Fig. 4(a) and Fig. 4(b), traditional trackers exhibit considerable drift when faced with multiple distractors. In contrast, RAMCT maintains accurate target localization by leveraging a spatially adaptive binary mask combined with context-aware regularization. This design dynamically suppresses background interference and emphasizes target-relevant regions, effectively minimizing false positives by reducing spurious correlations with non-target features.

In Fig. 4(c) and Fig. 4(d), other trackers fail to recover target positions following partial occlusions. By contrast, the proposed RAMCT leverages its GSVD-based region-adaptive iterative Tikhonov regularization to preserve target representation consistency and enhances feature stability during occlusion events. The advantages of multi-channel feature fusion are particularly evident in complex backgrounds (Fig. 4(e), "person_H_002"). While other trackers struggle with drift due to thermal crossover and intensity variations, our RAMCT ensures robust TIR tracking through optimized HOG-CNN feature complementarity and region-sparse constraints. Furthermore, as shown in Fig. 4(f), our RAMCT demonstrates exceptional tracking performance in high-speed motion scenarios. The dynamic discrepancy principle adjustment enables real time filter updates, ensuring accurate motion compensation through adaptive parameter refinement. As shown in Fig 4(d), this capability allows the proposed RAMCT to maintain precise localization even under rapid target displacement.

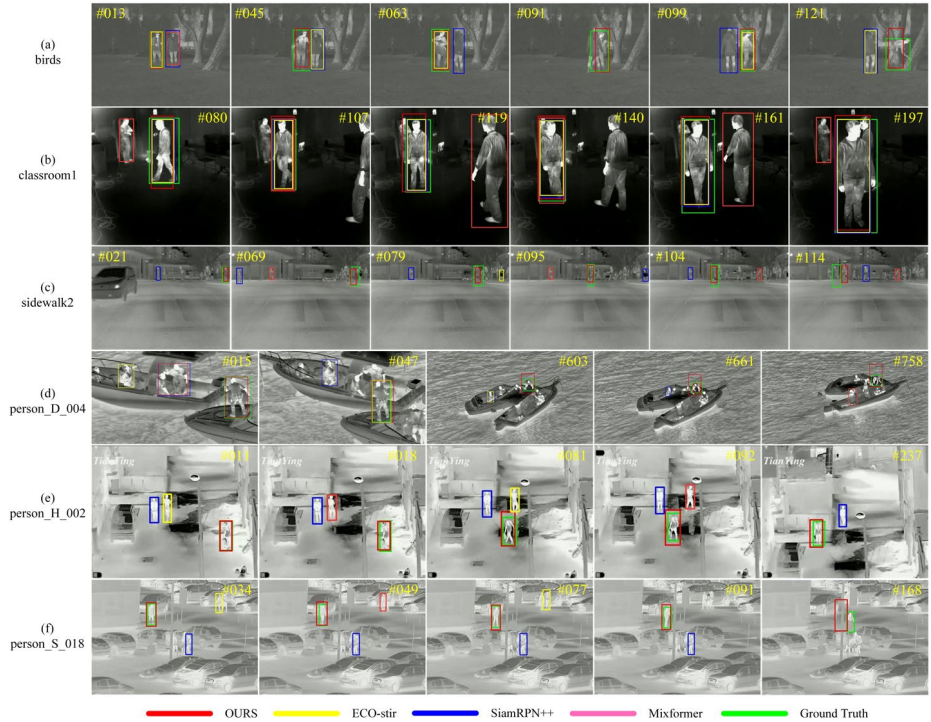

**Fig. 4.** Visualization results of the qualitative comparison experiments conducted on six challenging sequences in the PTB-TIR and LSOTB-TIR benchmarks.

## 5   Conclusions

In this study, we introduced a region-adaptive sparse CF tracker, RAMCT, to address the challenges of TIR target tracking. By integrating multi-channel feature optimization, GSVD-based region-adaptive iterative Tikhonov regularization, and an online optimization strategy with dynamic discrepancy-based parameter adjustment, our tracker effectively overcomes the limitations of traditional CF-based approaches, particularly in handling occlusion, background clutter, and target deformation. To evaluate the effectiveness of RAMCT, we conducted extensive experiments on four challenging benchmarks, and the results demonstrated that our tracker outperforms other state-of-the-art trackers, delivering better accuracy and robustness.